# ПРЯМА ТА ЗВОРОТНЯ ЗАДАЧИ ДИНАМІКИ ДЛЯ ТРИКОЛІСНОГО МОБІЛЬНОГО РОБОТА З ДВОМА ВЕДУЧИМИ КОЛЕСАМИ

**Волянський Р.С., к.т.н., доц.**
*КПІ ім. Ігоря Сікорського, кафедра автоматизації електромеханічних систем та електроприводу*

**Вступ.** В наш час мобільні роботи широко застосовуються при виконанні різних технологічних операцій у цілому ряді галузей народного господарства. Ці операції пов'язані з транспортуванням вантажів та обладнання, виконання робіт щодо визначення стану технічного об'єкту чи споруди, їх конструювання чи ремонту, виконання робіт з дослідження певної території та складання відповідних карт тощо. В останній час перелік операцій, які можуть виконувати мобільні роботи розширився поліцейськими та військовими операціями [1-3].

Очевидно, що від швидкості та точності руху таких роботів залежить безпека персоналу, який працює поруч, та час, який необхідний для виконання відповідних операцій. Тому виникає важлива задача дослідження та формування траєкторій руху мобільних роботів.

В наш час така задача розв'язується за допомогою методів та підходів сучасної теорії керування. В першу чергу до таких методів слід віднести методи інтелектуального керування, яке базується на використанні нечіткої логіки та машинного навчання [4-5]. Суттєвим недоліком таких методів є високій рівень суб'єктивізму з боку проектувальника відповідних систем керування та персоналу, який здійснює їх налагодження. Відповідно функціонування системи керування залежить від рівня знань та навичок конструкторсько-налагоджувального персоналу і може відбуватися у непередбачених на етапі проектування системи режимах у випадку недостатньої проробки алгоритмів керування. З іншого боку, використання класичних алгоритмів автоматичного керування для керування роботами не гарантує стійкості їх руху та бажаної швидкодії [6,7]. Тому на теперішній час найбільш придатними для керування мобільним роботами будемо вважати методи, які базуються на новітніх розробках в області теорії автоматичного керування, які базуються на методах оптимізації, адаптації та робастності і використанні теорії стійкості руху [8-10].

Використання цих методів дозволяє розглядати мобільний робот [11-15] як де-яку динамічну систему з декількома входами та виходами. Математичний опис такої динамічної системи [16-27] може бути використаний для аналізу та синтезу бажаних траєкторій руху шляхом розв'язання відповідних прямої та зворотної задач динаміки [28-32]. Тому створення математичної моделі мобільного робота є актуальною задачею, розв'язання якої дозволяє створити та дослідити системи керування роботом, які забезпечують рух по наперед-заданим бажаним траєкторіям.

**Мета роботи.** Розробка високо-формалізованої математичної моделі, триколісного мобільного робота з двома ведучими колесами, які приводяться до руху синхронними реактивними двигунами та використання цієї моделі для синтезу багатоканальної системи керування рухом робота.

**Матеріали і результати досліджень.** Для складання рівнянь руху робота розглянемо його розрахункову схему, яка наведена на рис.1

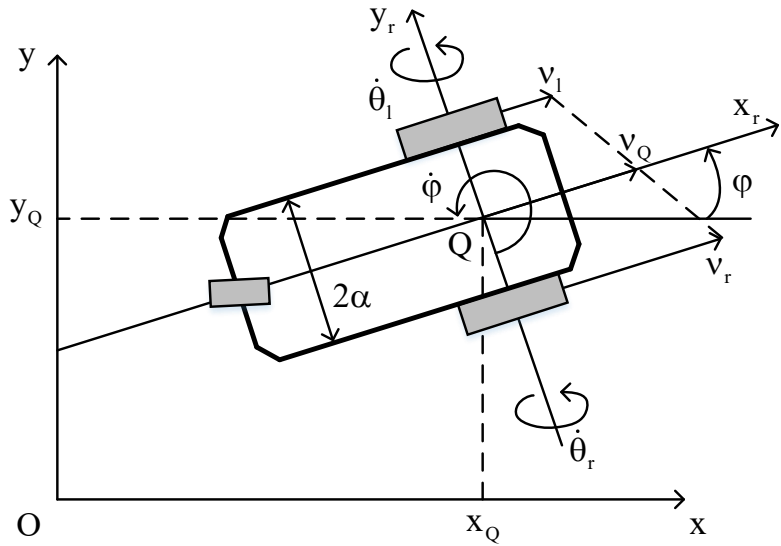

Рисунок 1 - Кінематична схема мобільного робота з диференціальним приводом

При розробці математичної моделі будемо користуватися наступними припущеннями:
- Колеса при русі не мають ефекту просковзування;
- Колеса розташовані перпендикулярно площині Оху, тому при рульовому керування відхилення відсутнє;
- Точка Q, що знаходиться посередині вісі приводів збігається з центром ваги робота.

Будемо вважати, що положення і орієнтація робота [1] в будь-який момент часу описується вектором $p = [x_Q, y_Q, \varphi]^T$, де $(x_Q, y_Q)$ – положення точки Q (центра осі приводних коліс) та $\varphi$ – орієнтація робота в глобальній системі координат, кутові швидкості лівого і правого коліс позначено $\theta_l$ і $\theta_r$ відповідно.

Приймаючи до уваги складність розглядаємої електромеханічної системи, яка складається з двох приводних двигунів, які обертають колеса та взаємовпливають один на одний через робототехнічну платформу, ми будемо використовувати для створення динамічної моделі лагранжевий формалізм. Відповідно до методу Лагранжа, рух будь-якої динамічної системи може бути описаний за допомогою рівнянь Лагранжа

$$\frac{d}{dt}\frac{\partial L}{\partial \dot{w}_j} - \frac{\partial L}{\partial w_j} + \frac{\partial T_e}{\partial \dot{w}_j} = \sum Q_j, \qquad (1)$$

де $w_j$ – узагальнена координата, L – лагранжіан розглядаємої динамічної системи, $T_e$ – функція Реллея, $Q_j$ – узагальнені зовнішні зусилля.

В якості узагальнених координат ми використовуємо лінійний та кутовий шлях, пройдений роботом.

В загальному випадку лагранжіан системи може бути визначений наступним чином

$$L = T_k - T_g, \qquad (2)$$

де кінетична енергія $T_к$ складається з енергій лінійного та кутового рухів платформи і коліс

$$T_k = 0.5 \sum m_i V_i^2 + 0.5 \sum J_i \omega_i^2, \qquad (3)$$

тут $m_i$ – маси коліс та платформи, $V_i$ – їх лінійні швидкості, $J_i$ – моменти інерції коліс та платформи відносно осей обертання, $\omega_i$ – їх кутові швидкості, $T_g$ – потенційна енергія робота, приймаючи до уваги, що рух відбувається у горизонтальній площині, ця енергія прирівнюється до нуля.

З врахуванням останнього твердження лагранжіан розглядаємого мобільного робота можна однозначно співставити з його кінетичною енергією

$$L = T_k \qquad (4)$$

та записати у розгорнутому вигляді наступним чином

$$L = (m_1 + 2m_k)(\dot{x}_Q^2 + \dot{y}_Q^2) + J_y(\omega_l^2 + \omega_r^2) - 2m_1 a(\dot{x}_Q \sin\varphi - \dot{y}_Q \cos\varphi)\dot{\varphi} + \\ + (m_1 a^2 + J_1 + 2m_k l^2 + 2J_{kz})\dot{\varphi}^2, \qquad (5)$$

тут приведений момент інерції коліс та двигунів визначається формулою

$$J_y = J_{ky} + n^2 J_{ry}. \qquad (6)$$

Підстановка виразу (5) до рівнянь (1) дозволяє записати наступні диференційні рівняння, які описують динаміку розглядаємої робототехнічної платформи наступним чином

$$m\dot{V} - m_1 a \omega^2 = \frac{M_1 + M_2}{r}; \quad J\dot{\omega} + m_1 a V \omega = \frac{M_1 - M_2}{r}, \qquad (7)$$

де $M_1$ та $M_2$ – електромагнітні моменти лівого та правого привідних двигунів, $a$ – напіввідстань між колесами,

$$m = m_1 + 2m_k + 2J_y/r^2; \quad J = J_1 + 2J_{kz} + (m - m_1)l^2 + m_1 a^2. \qquad (8)$$

Доповнивши рівняння (7) рівняннями кінематики та виконавши перехід до операторної форми, отримаємо такі рівняння

$$sx = V\cos\psi; \quad sy = V\sin\psi; \quad s\psi = \omega;$$
$$sV = a_{45}\omega^2 + m_4(M_1 + M_2); \quad s\omega = a_{54}V\omega + m_5(M_1 - M_2), \qquad (9)$$

де

$$a_{45} = m_1 a / m; \quad m_4 = 1/mr; \quad a_{54} = -m_1 a / J; \quad m_5 = 1/Jr. \qquad (10)$$

Система рівнянь (9) описує рух механічної частини мобільного робота в залежності від прикладених до його коліс моментів привідних двигунів.

Доповнимо цю модель рівнянням електричної рівноваги k-ої фази СРМ

$$U_k = i_k R_k + i_k \frac{\partial L_k(\theta, i_k)}{\partial \theta}\omega + L_k(\theta, i_k) s i_k, \qquad (11)$$

де $R_k$ – опір k-ої фази, $L_k$ – індуктивність k-ої фази, $i_k$ – струм k-ої фази, $\omega$ та $\theta$ – швидкість і кут обертання ротора відповідно,
рівнянням електромагнітного моменту, який створюється цією фазою

$$M_k = i_k^2 \frac{\partial L_k(\theta, i_k)}{\partial \theta} \qquad (12)$$

та виразом для сумарного електромагнітного моменту

$$M = \sum_{k=1}^{n} M_k \qquad (13)$$

Математична модель, складена на основі рівнянь (9), (11)-(13) може розглядатися як розв'язання прямої задачі динаміки.

Взагалі-то ця модель може бути використана і для розв'язання зворотної задачі динаміки. Однак, приймаючи до уваги, змінність структури моделі двигуна, викликану перемиканням його фаз, при розв'язанні зворотної задачі будемо вважати, що до складу системи керування входить швидкодіючий регулятор моменту. Введення такого контуру та організація в ньому ковзних режимів дозволяє гарантувати безінерційність контуру регулювання моменту і таким чином виключити із розгляду рівняння, які описують динаміку двигунів. Таким чином розв'язання зворотної задачі динаміки для розглядаємого робота може виконуватися виключно на основі рівнянь (9).

Послідовно розв'яжемо перші три рівняння (9) відносно лінійної та кутової швидкостей платформи

$$\psi = arctg \frac{sy}{sx}; \quad V = \frac{sx}{\cos\psi}; \quad \omega = s\psi = s \cdot arctg \frac{sy}{sx} = \frac{1}{1+\left(\frac{sy}{sx}\right)^2} \frac{s^2 y sx - sy s^2 x}{(sx)^2}. \quad (14)$$

За відомими швидкостями платформи знайдемо моменти, які мають бути розвинені двигунами, щоб забезпечити такі швидкості. Для знаходження цих моментів будемо розв'язувати останні два рівняння системи (9)

$$M_1 = \frac{sV - a_{45}\omega^2}{2m_4} - \frac{a_{54}V\omega - s\omega}{2m_5}; \quad M_2 = \frac{sV - a_{45}\omega^2}{2m_4} + \frac{a_{54}V\omega - s\omega}{2m_5}; \qquad (15)$$

Моменти (15) будемо розглядати, як сигнали завдання для контурів керування моментами кожного двигуна.

Вирази (14)-(15) є шуканим розв'язанням зворотної задачі динаміки.

Використання цих виразів дозволяє побудувати теоретично безінерційну двоканальну систему керування рухом мобільного робота, яка здатна відтворювати задану траєкторію руху. Функціональна схема такої системи керування наведена на рис.2.

Наведена структура має два суттєвих недоліки:

1. Вона є розімкненою відносно координат робота, що може призвести до неточного відпрацювання бажаної траєкторії у випадку коли параметри, використовувані при налаштуванні блоку формувача моменту відрізняються від реальних параметрів робота. Таке може статися внаслідок зміни умов роботи, впливу оточуючого середовища, тощо.

2. Задаючи моменти, які формуються в цій структурі можуть досягати великих значень, які неможна відтворити фізично, внаслідок невірного завдання бажаної траєкторії руху.

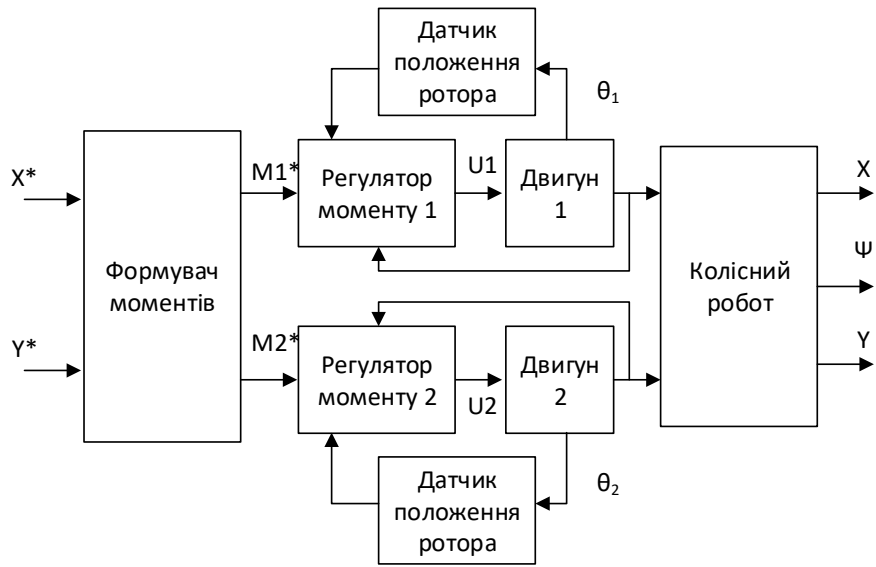

Рисунок 2 – Функціональна схема системи керування положенням мобільного робота

Усунути ці недоліки можна шляхом замикання системи керування за положенням робота (рис.3)

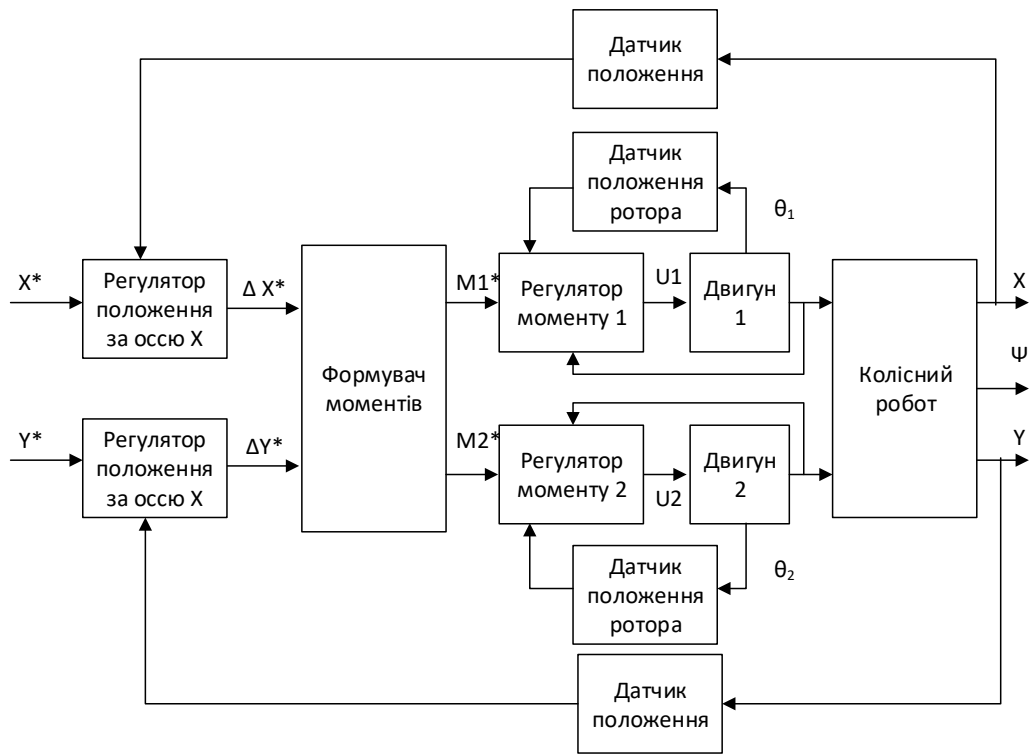

Рисунок 3 – Функціональна схема замкненої системи керування положенням мобільного робота

В якості датчиків положення платформи робота можуть використовуватися різного роду дальноміри та оптичні системи, побудовані на основі розпізнавання образів, якщо рух відбувається у відомому закритому приміщенні, або датчики системи GPS у випадку руху назовні. В останньому випадку необхідно потурбуватися про підвищення точності отримання сигналів позиціювання.

Окрім порівняння бажаної та поточної координати робота регулятори положення на вході системи забезпечують формування фізично-реалізуємих моментів двигуні в та обмежують їх значення

На рис.4 наведені перехідні процеси відпрацювання складної траєкторії такою системою

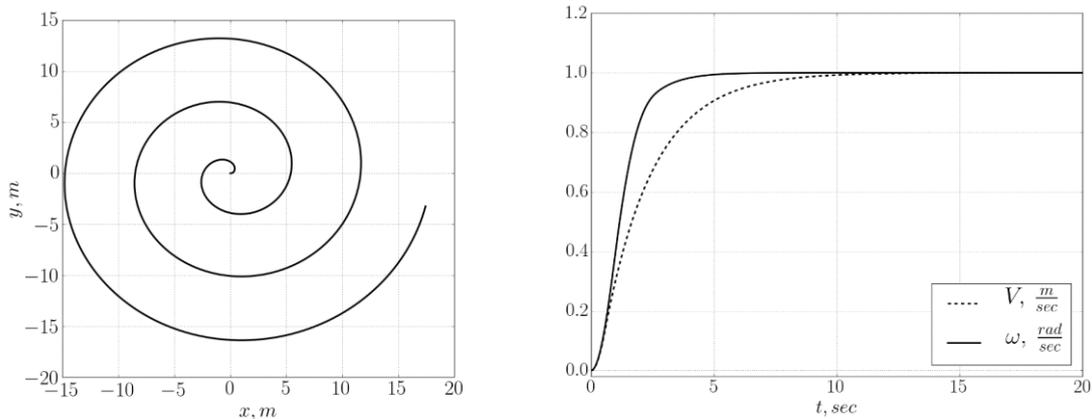

Рисунок 4 – Відпрацювання складної траєкторії руху в синтезованій замкненій системи керування положенням мобільного робота

Аналіз наведених графіків показує високу якість відтворення заданої траєкторії руху та асимптотичну стійкість синтезованої системи керування.

**Висновки.** Використання формалізованого підходу до створення математичних моделей, який базується на рівняння Лагранжа, дозволяє створювати математичні моделі складних багатоканальних нелінійних електромеханічних систем, розв'язуючи тим самим пряму задачу динаміки та визначаючи траєкторії руху динамічного об'єкта за поданими сигналами керування. Ці моделі можуть бути спрощені шляхом використання методів побудови робастних систем керування які працюють у ковзних режимах. Завдяки виникненню цих режимів вдається компенсувати внутрішні нелінійності об'єкта керування та перетворити його на безінерційний динамічний об'єкт.

В свою чергу, розв'язання зворотної задачі динаміки дозволяє визначити структуру та параметри системи керування, яка забезпечує рух динамічного об'єкта заданими траєкторіями. Теоретично, знайдені в ході такого розв'язання регулятори дозволяють побудувати безінерційну систему керування. Однак в практичних випадках внаслідок інерційності об'єкта керування це може призвести до появи нескінчено великих сигналів керування, які неможливо фізично реалізувати. Тому до відповідних алгоритмів керування мають бути додані інтегруючі складові, які дозволяють створити замкнену систему керування з бажаною інерційністю.